\documentclass[conference, compsocconf]{IEEEtran}
\usepackage[cmex10]{amsmath}
\usepackage{url}
\usepackage{tikz-qtree}
\usepackage{graphicx}
\usepackage{caption}
\usepackage{fixltx2e}
\usepackage{float}
\usepackage{mathtools}
\usepackage{amssymb}
\usepackage{pst-tree}
\usepackage{pst-node} 
\usepackage{pstricks} 
\usepackage{pstricks-add}
\usepackage{graphicx}
\usepackage[utf8]{vietnam}

\DeclareMathOperator*{\yy}{\mathbf y}

\DeclareMathOperator*{\xx}{\mathbf x}

\begin{document}

\captionsenglish

\title{An Empirical Study of Discriminative Sequence Labeling Models
  for Vietnamese Text Processing}

\author{
\IEEEauthorblockN{
Phuong Le-Hong\IEEEauthorrefmark{1}, Minh Pham Quang
Nhat\IEEEauthorrefmark{2}, Thai-Hoang Pham\IEEEauthorrefmark{3}, Tuan-Anh
Tran\IEEEauthorrefmark{4}, Dang-Minh Nguyen\IEEEauthorrefmark{4}}
\IEEEauthorblockA{\IEEEauthorrefmark{1}College of Science, Vietnam National University, Hanoi, Vietnam}
\IEEEauthorblockA{\IEEEauthorrefmark{2}FPT Technology Research
  Institute, FPT University, Hanoi, Vietnam}
\IEEEauthorblockA{\IEEEauthorrefmark{3}Alt Vietnam}
\IEEEauthorblockA{\IEEEauthorrefmark{4}FPT Technology Innovation, FPT Corp., Vietnam}
}

\maketitle

\begin{abstract}
  This paper presents an empirical study of two widely-used sequence
  prediction models, Conditional Random Fields (CRFs) and Long
  Short-Term Memory Networks (LSTMs), on two fundamental tasks for
  Vietnamese text processing, including part-of-speech tagging and
  named entity recognition. We show that a strong lower bound for
  labeling accuracy can be obtained by relying only on simple
  word-based features with minimal hand-crafted feature engineering,
  of 90.65\% and 86.03\% performance scores on the standard test
  sets for the two tasks respectively. In particular, we demonstrate
  empirically the surprising efficiency of word embeddings in both of
  the two tasks, with both of the two models. We point out that the
  state-of-the-art LSTMs model does not always outperform
  significantly the traditional CRFs model, especially on
  moderate-sized data sets. Finally, we give some suggestions and
  discussions for efficient use of sequence labeling models in
  practical applications.
\end{abstract}

\IEEEpeerreviewmaketitle

\section{Introduction}

Many datasets, such as text collections and genetic databases, consist
of sequences of distinct values. For applications that use such datasets, we often need to predict the sequence of labels given an observation sequence. In sequence prediction problems,
we attempt to predict elements of 
a sequence on the basis of the preceding elements.  Many statistical
sequence models have been developed for sequence prediction, for
example hidden Markov models (HMM)~\cite{Rabiner:1989,Welch:2003},
maximum entropy Markov models (MEMMs)~\cite{McCallum:2000},
conditional random fields (CRFs)~\cite{Lafferty:2001} or recurrent
neural nets (RNNs)~\cite{Hochreiter:1997}. These are all
powerful probabilistic tools for modeling sequential data and have
been applied to many text-related tasks such as part-of-speech
tagging, named entity recognition, text segmentation and information
extraction. These models also support applications in bioinformatics
such as expressed sequence tag finding and gene discovery.

In this paper, we present an empirical study of two prevalent
discriminative sequence labeling models, CRFs and LSTMs, on two
fundamental problems of text processing, namely part-of-speech tagging
and named entity recognition. Experiments are carefully designed, carried out and analyzed on standard Vietnamese data
sets. The main findings of this work are as follows. First, we show
that we can obtain a strong performance lower bound for both
part-of-speech (POS) tagging and named entity recognition (NER) by using only
simple and word-based features, both with CRFs and LSTMs. For
POS tagging, we achieve a test accuracy of 90.65\% by
using only word identities, word shapes and word embedding features on
sentences less than 25 tokens. For NER with the
same feature set and sentence length, we obtain about 86.03\% of
F1-score. Second, we show that word embeddings
are very effective and beneficial for both of the two tasks. They help
improve POS tagging accuracy significantly by about 4.0\%
when using with LSTMs or 1.35\% when using with CRFs. Word embeddings
are even more beneficial for NER -- they help improve recognition
performance by more than 5\% in both of the models.  Third, we show
that although the LSTMs model slightly outperforms CRFs in terms of
accuracy, the gap is relatively small, especially on moderate-sized data
sets, with the cost of much longer training time. Finally, this paper
gives some suggestions for efficient use of sequence labeling models
in practical applications.

The remainder of this paper is structured as
follows. Section~\ref{sec:methodology} presents the adopted
methodology. Section~\ref{sec:experiments} describes detailed settings
and experimental results. Section~\ref{sec:conclusion} gives
discussions and findings. Section~\ref{sec:relatedWork} presents
related work. Finally, Section~\ref{sec:conclusion}
concludes the paper.

\section{Methodology}
\label{sec:methodology}

\subsection{Fundamental Tasks}
This subsection gives a brief description of two fundamental sequence
learning tasks investigated in this study, part-of-speech tagging and
named entity recognition. 

\subsubsection{Part-of-Speech Tagging}

POS tagging is a typical sequence prediction task, where we are
interested in building a model that reads text in some languages and
assigns a part-of-speech to each token (word), such as noun, verb,
adjective. In general, POS taggers in computational applications use
more fine-grained POS tags like common noun or proper noun. For
example, each word of the following English sentence is tagged with
its most likely correct part-of-speech:

\textit{Profits/\texttt{N} soared/\texttt{V} at/\texttt{P} Boeing/\texttt{N} Co./\texttt{N} ,/\texttt{,} easily/\texttt{ADV} topping/\texttt{V}
forecasts/\texttt{N} on/\texttt{P} Wall/\texttt{N} Street/\texttt{N} ,/, as/\texttt{P} their/\texttt{POSS} CEO/\texttt{N}
Alan/\texttt{N} Mulally/\texttt{N} announced/\texttt{V} first/\texttt{ADJ} quarter/\texttt{N} results/\texttt{N} ./\texttt{.}}

\noindent where the tags \texttt{N}, \texttt{V}, \texttt{P}, \texttt{ADV},
\texttt{ADJ} denotes a noun, a verb, a preposition, an adverb, an
adjective, respectively. 

\subsubsection{Named Entity Recognition}

Named entity recognition, also known as entity identification is a
subtask of information extraction that aims to locate and classify
elements in texts into pre-defined categories such as the names of
persons, organizations, locations and so on.

For example, the named entities extracted from the same English
sentence above are as follows:

\textit{Profits soared at [\texttt{Organization} Boeing Co.], easily
  topping forecasts on [\texttt{Location} Wall Street], as their CEO
  [\texttt{Person} Alan Mulally] announced first quarter results.}

In this example, an organization name, a location and a person name have been
detected and classified. Actually, NER can be formalized as a sequence tagging problem,
where each token is tagged with a specific tag, for example:

\textit{Profits/\texttt{O} soared/\texttt{O} at/\texttt{O}
  Boeing/\texttt{B-ORG} Co./\texttt{I-ORG} ,/\texttt{O} easily/\texttt{O}
  topping/\texttt{O} forecasts/\texttt{O} on/\texttt{O}
  Wall/\texttt{B-LOC} Street/\texttt{I-LOC} ,/\texttt{O} as/\texttt{O}
  their/\texttt{O} CEO/\texttt{O} Alan/\texttt{B-PER}
  Mulally/\texttt{I-PER} announced/\texttt{O} first/\texttt{O}
  quarter/\texttt{O} results/\texttt{O} ./\texttt{O}}

Here, the tag \texttt{O} means ``no entity'', the tags \texttt{B-ORG} and
\texttt{I-ORG} mean ``begin organization'' and ``in organization''
respectively; similarly, the tags \texttt{B-LOC} and \texttt{I-LOC} mean
``begin location'' and ``in location'' respectively, and so on.

\subsection{Discriminative Models}

In this subsection, we give a brief description of two discriminative
sequence models used in this study, including Conditional Random Fields (CRFs), and Long Short-Term
Memory Recurrent Neural Networks (LSTMs).

\subsubsection{Conditional Random Fields}

Conditional Random Fields (CRF)~\cite{Lafferty:2001} is a discriminative probabilistic framework,
which directly model conditional probabilities of a tag sequence given
a word sequence. Formally, in CRF, the
conditional probability of a tag sequence $\yy = (y_1, y_2, \dots, 
y_T)$, given a word sequence $\xx = (x_1, x_2, \dots, x_T)$ is defined as follow.
\begin{equation*}
  P(\yy|\xx) \propto 
  \exp( \sum_{j} \lambda_j t_j (y_{i-1}, y_i, \xx, i) + \sum_{k} \mu_k s_k (y_i, \xx, i) )
\end{equation*}
where $t_j (y_{i-1}, y_i, \xx, i)$ is a transition feature function of
the entire observation sequence and the labels at the position $i$ and
$i-1$ in the label sequence; $s_k (y_i, \xx, i)$ is a state feature
function of the label at the position $i$ and the observation
sequence; $\lambda_j$ and $\mu_k$ are parameters to be estimated from
training data. 

We can simplify the notations by writing $s(y_i, \xx, i) = s(y_{i-1},\xx, i)$ and
\begin{equation*}
F_j(\yy,\xx) = \sum_{i=1}^n f_j(y_{i-1}, y_i, \xx, i)
\end{equation*}
where each $f_j(y_{i-1}, y_i,\xx,i)$ is either a state function $s(y_{i-1}, y_i, \xx)$
or a transition function $t(y_{i-1},y_i,\xx,i)$. By using this
notation, we can write the conditional probability as follows:
\begin{equation*}
  \label{eq:condProb}
P(\yy|\xx,\lambda) = \frac{1}{Z(\xx)} \exp( \sum_j \lambda_j F_j(\yy, \xx) )
\end{equation*}
$Z(\xx)$ is a normalization factor.

The parameters in CRF can be estimated by maximizing log-likelihood objective function:
\begin{equation*}
L(\lambda) = \sum_k [ \log\frac{1}{Z(\xx^{(k)})} + \sum_j \lambda_j
  F_j(\mathbf{y}^{(k)}, \mathbf{x}^{(k)}) ]
\end{equation*}
Parameter estimation in CRF can be done by using iterative scaling
algorithms~\cite{Darroch:1972,Lafferty:2001,Goodman:2002}  or
gradient-based methods~\cite{Nocedal:2006}.

\subsubsection{Long Short-Term Memory Networks}

Recurrent Neural Networks (RNNs) have recently been widely used for sequence
labelling because they can directly represent sequential structures
such as word sequences, sounds and time series data. For this reason,
there is a rapidly growing interest in using RNNs for practical
applications as an efficient method to map input sequences to output
sequences. They are computationally more powerful and biologically
more plausible than other adaptive approaches such as Hidden Markov
Models (no continuous internal states), Feed-Forward Neural Networks (FFNN)
and Support Vector Machines (no internal states at all).\footnote{Many
  interesting details of RNNs are available online at
  \url{http://people.idsia.ch/~juergen/rnn.html}} Traditional RNNs of
the 1990s could not learn to look far back into the past because of
the vanishing or exploding gradient problems. A feedback network
called Long Short-Term Memory (LSTM)~\cite{Hochreiter:97} was proposed
to overcome these problems.

We represent the word sequence of a sentence with a bidirectional
LSTM~\cite{Graves:13}. The LSTM unit at the $t$-th word consists of a
collection of multi-dimensional vectors, including an input gate
$i_t$, a forget gate $f_t$, an output gate $o_t$, a memory cell $c_t$,
and a hidden state $h_t$. The unit takes as input a $d$-dimensional
input vector $x_t$, the previous hidden state $h_{t-1}$, the previous
memory cell $c_{t-1}$, and calculates the new vectors using the
following six equations:
\begin{align*}
  i_t &= \sigma(W^i  x_t + U^i h_{t-1} + b^i)  \\
  f_t &= \sigma(W^f x_t + U^f h_{t-1} + b^f)  \\
  o_t &= \sigma(W^o x_t + U^o h_{t-1} + b^o)  \\
  u_t &= \tanh(W^u x_t + U^u h_{t-1} + b^u)  \\
  c_t &= i_t \cdot u_t + f_t \cdot c_{t-1} \\
  h_t &= o_t \cdot \tanh(c_t),
\end{align*}
where $\sigma$ denotes the logistic function, the dot product 
denotes the element-wise multiplication of vectors, $W$ and $U$ are
weight matrices and $b$ are bias vectors. The LSTM unit at $t$-th word
receives the corresponding word embedding as input vector
$x_t$. Since the LSTM is bidirectional, we concatenate the hidden
state vectors of the two directions' LSTM units corresponding to each
word as its output vector and pass it to the subsequent layer.

\section{Experiments}
\label{sec:experiments}

\subsection{Datasets}

\subsubsection{Part-of-Speech Tagging}

We perform experiments on Vietnamese part-of-speech tagging using the
standard part-of-speech tagged corpus of the VLSP
project.\footnote{\url{https://vlsp.hpda.vn/demo/?page=home}}. This corpus contains
10,165 manually tagged sentences where the training set contains 9,000
sentences and the test set contains 1,165 sentences. The tagset has 21
different tags. Further details of the corpus are described
in~\cite{VTB:2009}. 

\subsubsection{Named Entity Recognition}

For experiments on named entity recognition, we use the standard NER
corpus developed by the Vietnamese Language and Speech
Processing\footnote{\url{http://vlsp.org.vn/}} community in late 
2016. Similar to the CoNLL 2003 NER corpus for English, four named
entity types are considered, including persons (PER), organizations
(ORG), locations (LOC), and miscellaneous entities (MISC). The data
are collected from electronic newspapers published on the
web. Table~\ref{tab:ner} shows the quantity of named entity annotated
in the training set and the test set.
\begin{table}
  \center
  \caption{Statistics of named entities in the VLSP corpus}
  \label{tab:ner}
  \begin{tabular}{|l|r|r|}
    \hline 
    \textbf{Entity Types} & \textbf{Training Set} & \textbf{Test Set}
    \\ 
    \hline
    \hline 
    Location & 6,247 & 1,379 \\ 
    \hline 
    Organization & 1,213 & 274 \\ 
    \hline 
    Person & 7,480 & 1,294 \\ 
    \hline 
    Miscellaneous names & 282 & 49 \\ 
    \hline 
    All & 15,222 & 2,996 \\ 
    \hline 
  \end{tabular}
\end{table}

\subsection{Feature Sets}

\subsubsection{Word Identities}

The first basic feature set contains only word occurrence information
extracted from the training set. All words having an occurrence
frequency above a minimum threshold are kept in a vocabulary
$\mathcal V$. Each word can be represented by an one-hot sparse
vector of size $|\mathcal V|$.

\subsubsection{Word Shapes}

In addition to word identities, word shapes have been shown to be
important features for improving prediction ability, especially 
for unknown or rare words. Common word shape features used in our
experiments are shown in Table~\ref{tab:wordForms}. We used regular expressions to extract those word shape features.

\begin{table}[t]
  \caption{Word shape features}
  \label{tab:wordForms}
  \centering
  \begin{tabular}{|l | l |}
    \hline
    \textbf{Feature} & \textbf{Example} \\ \hline \hline
    is lower word  & ``\textit{tỉnh}'' \\ \hline
    is capitalized word &``\textit{Tổng\_cục}'' \\ \hline
    contains all capitalized letters (allcaps) & ``\textit{UBND}'' \\ \hline
    is name -- consecutive syllables are capitalized &
                                                          ``\textit{Hà\_Nội}'', ``\textit{Buôn\_Mê\_Thuột}'' \\ \hline
    is mixed case letters & ``\textit{iPhone}'' \\ \hline
    is capitalized letter with period & ``\textit{H.}'', ``\textit{Th.}'',
                                        ``\textit{U.S.}'' \\ \hline
    contains hyphen & ``\textit{H-P}'' \\ \hline
    is number & ``\textit{100}'' \\ \hline
    is date & ``\textit{20-10-1980}'', ``\textit{10/10}'' \\ \hline
  \end{tabular}
\end{table}

\subsubsection{Word Embeddings}

Word embeddings are low-dimensional distributed representation of
words. Each word embedding is a real-valued vector of $d$ dimensions
where $d$ is much smaller than $|\mathcal V|$ of its one-hot sparse
representation. Distributed word representations have been shown very
useful for many natural language processing tasks. Many
state-of-the-art language processing models are now employing word or
character embeddings. In particular, some previous works have
also integrated Vietnamese word embeddings to improve performance~\cite{Le:2015c,Vu:2015}.

To create distributed word representations, we use a dataset
consisting of 7.3GB of text from 2 million articles collected
through a Vietnamese news portal.\footnote{\url{http://www.baomoi.com}} 
The text is first normalized to lower case and all special characters are
removed except these common symbols: the comma, the semicolon, the
colon, the full stop and the percentage sign. All numeral sequences
are replaced with the special token \textless number\textgreater,\ so
that correlations between certain words and numbers are correctly
recognized by the neural network or the log-bilinear regression model.

Each word in the Vietnamese language may consist of more than one
syllables with spaces in between, which could be regarded as multiple
words by the unsupervised models. Hence it is necessary to replace the
spaces within each word with underscores to create full word
tokens. The tokenization process follows the method described
in~\cite{Le:2008a}. After removal of special characters and tokenization, the articles add
up to $969$ million word tokens, spanning a vocabulary of $1.5$ million unique
tokens. We train the unsupervised models with the full vocabulary to obtain
the representation vectors, and then prune the collection of word
vectors to the $65,000$ most frequent words, excluding special symbols and
the token \textless number\textgreater \, representing numeral sequences.
We train the Mikolov's continuous Skip-gram model using the neural
network and source code introduced in~\cite{Mikolov:2013b}. The
continuous skip-gram model itself is described in details
in~\cite{Mikolov:2013a}. Each word is represented by a real-valued
vector of 25 dimensions.

In both CRF and LSTM models, we use three kinds of features mentioned above. In the CRF model, we represent word identities, word shapes as binary features. Each dimension of a word-embedding vector is a feature and its value is the feature value. In the LSTM model, we use word shape and word embedding features as additional dimensions in vector representation for each word. Thus,  each word in the LSTM model is represented by a vector of size $|\mathcal V| + 34$ ($|\mathcal V|$ is the vocabulary size; we use 9 word shape features and 25-dimension word-embedding vectors).

\subsection{Evaluation Method}

For the POS task, our system is evaluated by the tagging accuracy on
the corresponding data sets. The accuracy is the ratio of number of tokens which
are correctly tagged divided by the total number of tokens in the test
set. For the NER task, the performance of our system is measured with
$F_{1}$ score: $F_1 = 2*P*R / (P+R)$. Precision ($P$) is the
percentage of named entities found by the learning system, which are
correct predictions. Recall ($R$) is the percentage of named entities present in
the corpus that are found by the system. A named entity is correct
only if it is an exact match of the corresponding entity in the data
file. The performance of our system is evaluated by the automatic
evaluation script of the CoNLL 2003 shared
task.\footnote{\url{http://www.cnts.ua.ac.be/conll2003/ner/}}

\subsection{Experimental Settings}

In the experiments, we fix the minimum frequency threshold for
features as 5. In other words, all words or tags which do not appear
at least 5 times in the training corpus are considered unknown. 
In our experiments with the CRF model, we adopted
CRFsuite~\cite{CRFsuite}, an implementation of linear-chain
(first-order Markov) CRF. That toolkit allows us to easily incorporate
both binary and numeric features such as word embedding features. We
use default setting of CRFsuite in which the training algorithm is
L-BFGS~\cite{nocedal1980updating} and L2 regularization is used. The coefficient for L2 regularization is $1.0$.

The recurrent neural networks all have one bidirectional recurrent
layer of different numbers of units whose activation function is
$\tanh$. The output layer uses softmax activation function as
usual. The multiclass cross entropy loss function is selected. The
network is trained by using the stochastic gradient descent
optimization algorithm with learning rate fixed at 0.01. The Xavier
initilizer is used for parameter initialization~\cite{Xavier:2010}. We
use early stopping when training the network to help avoid overfitting
and remove the need to manually set the number of training epoch. The
training terminates either if the training score does not improve for
three consecutive epoches or if the number of epoches reaches 400.

\subsection{Results}

This subsections presents experimental results of the models on the 
two tasks. We first present results of part-of-speech tagging, and then 
those of named entity recognition.

\subsubsection{Part-of-Speech Tagging}

In the first experiment, we train and compare performance of sequence
models on sentences of length not greater than 20 tokens. There are
4,879 training sentences and 570 test sentences. The vocabulary size
is 1,630.

We train different LSTMs with varying number hidden units in the range from 32
to 200. Table~\ref{tab:pos-lstm-1} shows their performance on the feature set
\{\textit{word identity}, \textit{word shapes}\}. We see that the larger number
of hidden units is, the better result the tagger can achieve on the test set. The
LSTMs tagger achieves 85.98\% of accuracy on the test set when the network has
200 hidden units.

\begin{table}[t]
  \caption{Performance of LSTMs for PoS tagging using word identities
    and word shapes}
  \label{tab:pos-lstm-1}
  \centering
  \begin{tabular}{| c | c | c | c | c |}
    \hline
    \textbf{Hidden Units}
    & \multicolumn{2}{c|}{\textbf{Test Acc.}}
    & \multicolumn{2}{c|}{\textbf{Training Acc.}} \\ \hline 
    & $n \leq 20$ & $n \leq 25$ & $n \leq 20$ & $n \leq 25$ \\
    \hline
    \hline
    16  &84.85 &86.45&94.91&96.40\\ \hline 
    32	&83.41&83.89&94.43 &93.18 \\ \hline
    64	&83.82&85.93&96.04&96.23 \\ \hline
    100	&83.49&85.53&93.93&95.74 \\ \hline
    128	&84.67&86.75&97.49&97.52 \\ \hline
    150	&85.84&86.33&97.49&97.27 \\ \hline
    200	&\textbf{85.98}&\textbf{87.46}&97.27&97.46 \\ \hline
  \end{tabular}
\end{table}

In the second experiment, we add word embeddings as features to the
LSTMs model to see whether they are helpful or not. Table~\ref{tab:pos-lstm-2} shows their performance
on the feature set \{\textit{word identity}, \textit{word shapes},
\textit{word embeddings}\}.

\begin{table}[t]
  \caption{Performance of LSTMs for PoS tagging using word identities,
    word shapes and word embeddings}
  \label{tab:pos-lstm-2}
  \centering
  \begin{tabular}{| c | c | c | c | c |}
    \hline
    \textbf{Hidden Units} & \multicolumn{2}{c|}{\textbf{Test Acc.}}
    & \multicolumn{2}{c|}{\textbf{Training Acc.}} \\ \hline
    & $n \leq 20$ & $n \leq 25$ & $n \leq 20$ & $n \leq 25$ \\
    \hline
    \hline
    16	&87.02 & 88.73 &96.60 & 97.79 \\ \hline
    32	&88.44 & 86.39 &98.46 & 93.41 \\ \hline
    64	&88.29 & 88.56 &99.06 & 96.55 \\ \hline
    100 &89.65 & 90.24 &99.21 & 98.84 \\ \hline
    128	&89.24 & 89.31 &98.79 & 98.26 \\ \hline 
    150	&89.83 & 89.65 &99.33 & 99.55 \\ \hline
    200	&\textbf{89.92} &\textbf{90.65} &99.71 &99.47 \\ \hline
  \end{tabular}
\end{table}

It is surprising that word embeddings helps improve the accuracy of the
tagger significantly. With the same training parameters, we are able
to boost the accuracy on the test set from 85.98\% to 89.92\%. This
result demonstrates that in LSTMs, it is beneficial to
combine both discrete features and continuous features to build a
better tagger.

\begin{table}[t]
  \caption{Performance of the CRF model for PoS tagging with two feature sets}
  \label{tab:pos-crf-1}
  \centering
  \begin{tabular}{| p{3.3cm} | c | c | c | c |}
    \hline
    \textbf{Feature set} & \multicolumn{2}{c|}{\textbf{Test Acc.}}
    & \multicolumn{2}{c|}{\textbf{Training Acc.}} \\ \hline
    & $n \leq 20$ & $n \leq 25$ & $n \leq 20$ & $n \leq 25$ \\
    \hline
    \hline
    \{Word identities, word shapes\} & 87.62 & 88.93 & 90.75 & 91.56 \\ \hline
    \{Word identities, word shapes, word embeddings\} &  \textbf{88.97} & \textbf{90.26} & 91.64 & 92.29 \\
    \hline
  \end{tabular}
\end{table}

Table~\ref{tab:pos-crf-1} shows the results of the CRF model with two
feature sets: 1) \{\textit{word identity}, \textit{word shapes}\}; and
2) \{\textit{word identity}, \textit{word shapes}, \textit{word
  embeddings}\}. The table indicates that incorporating word embedding
features helps to improves the accuracy of the CRF model 1.35\%
from 87.62\% to 88.97\%. The CRF model outperformed LSTMs when we do
not use word embedding features. However, its accuracy is lower than
that of LSTMs when word embedding features are incorporated.

In the third experiment, we enlarge the data set by considering longer
sentences. We train and compare sequence models on sentences not
longer than 25 tokens. With this length, there are 6,221 training
sentences and 737 test sentences in the standard data set. The
vocabulary now contains 2,197 different words. The LSTM tagger
achieves an accuracy of 90.65\% on the test set, significantly better
than its performance on the 20-token data set. Similar to the LSTM
models, the performances of CRFs model are better than those on
shorter sentences. This can be explained by the fact that the more
training data are available, the greater number of patterns the models
can learn. These results also confirm the effectiveness of word
embedding features for the CRF model.

We observe that performance of the CRF model is slightly worse
than that of LSTMs model, both on the test sets and on the training
sets. In addition, the LSTMs model has a very good
memorization capacity -- its accuracy on the training set is nearly
perfect on long sentences, especially when the number of hidden units
in use is large enough. 

\subsubsection{Named Entity Recognition}
Similar to PoS tagging experiments, we evaluate NER methods on
sentences of length not greater than 20 and on sentences of length not
greater than 25. In the former experiment, there are 8,968 training
sentences and 1,355 test sentences; the vocabulary size is 2,525. In
the latter experiment, there are 11,436 training sentences and 1,787
test sentences; the vocabulary size is 3,368. 

\begin{table}[t]
  \caption{Performance of the LSTM model for NER  with three feature
    sets on the test set (sentences not longer than 20 tokens)} 
  \label{tab:ner-lstm-1}
  \centering
  \begin{tabular}{| l | c | c | c |}
    \hline
    \textbf{Feature Set} & \textbf{Precision} & \textbf{Recall} & \textbf{F1}\\ \hline
    \hline
    (1) = Word identities, word shapes	& 77.29	& 70.17 & 77.23\\ \hline
    (2) = (1) + PoS tags & 78.92	& 82.85 & 80.68\\ \hline
    (3) = (2) + word embeddings & 85.29 & 86.77 & \textbf{86.03} \\ \hline
  \end{tabular}
\end{table}

\begin{table}[t]
  \caption{Performance of the LSTM model for NER  with three feature
    sets on the test set (sentences not longer than 25 tokens)} 
  \label{tab:ner-lstm-2}
  \centering
  \begin{tabular}{| l | c | c | c |}
    \hline
    \textbf{Feature Set} & \textbf{Precision} & \textbf{Recall} & \textbf{F1}\\ \hline
    \hline
    (1) = Word identities, word shapes	&  78.30 & 76.64  & 77.46 \\ \hline
    (2) = (1) + PoS tags & 81.27	& 82.53 &81.90 \\ \hline
    (3) = (2) + word embeddings &84.42  &87.36  &\textbf{85.86}  \\ \hline
  \end{tabular}
\end{table}

\begin{table}[t]
  \caption{Performance of the CRF model for NER  with three feature
    sets on the test set (sentences not longer than 20 tokens)} 
  \label{tab:ner-crf-1}
  \centering
  \begin{tabular}{| l | c | c | c |}
    \hline
    \textbf{Feature Set} & \textbf{Precision} & \textbf{Recall} & \textbf{F1}\\ \hline
    \hline
    (1) = Word identities, word shapes	& 77.78	& 70.55 & 73.99\\ \hline
    (2) = (1) + PoS tags & 78.20	& 77.95 & 78.08\\ \hline
    (3) = (2) + word embeddings & 85.74 & 84.25 & \textbf{85.00} \\ \hline
  \end{tabular}
\end{table}

\begin{table}[h]
  \caption{Performance of the CRF model for NER  with three feature
    sets on the test set (sentences not longer than 25 tokens)} 
  \label{tab:ner-crf-2}
  \centering
  \begin{tabular}{| l | c | c | c |}
    \hline
    \textbf{Feature Set} & \textbf{Precision} & \textbf{Recall} & \textbf{F1}\\ \hline
    \hline
    (1) = Word identities, word shapes	& 79.19	& 73.84 & 76.42\\ \hline
    (2) = (1) + PoS tags & 81.11	& 80.41 & 80.76\\ \hline
    (3) = (2) + word embeddings & 86.48 & 85.23 & \textbf{85.85} \\ \hline
  \end{tabular}
\end{table}

Table~\ref{tab:ner-lstm-1} and Table~\ref{tab:ner-lstm-2} shows the
performance of the LSTM models on sentences not longer than 20 and 25
tokens, respectively. Table~\ref{tab:ner-crf-1} and
Table~\ref{tab:ner-crf-2} shows the experimental results of the CRF
model on the same data sets and feature sets as those of LSTM
experiments. The results indicated the effectiveness of PoS tag and
word embedding features. While using PoS tag features mainly improved
recall using word embedding features helped to improve both precision
and recall. It can be explained that PoS features and especially word
embedding features can better capture semantic relationship between
words.

We see that the LSTM model is slightly better than the CRF model
on short sentences; while the two models perform similarly on longer
sentences. The best F-score of the LSTM model is about 86.03\%, which is not very far
below the state-of-the-art NER result on this data set, despite of the
minimal simplicity of the features in use.

\section{Discussion}
\label{sec:discussion}
Discriminative sequence labeling models such as CRF or LSTMs models
have been used for Vietnamese text processing tasks such as PoS
tagging or named-entity recognition (NER). In this work, we compare
the LSTMs model and the CRF model in two Vietnamese text processing
tasks. In our understanding, our work is the first empirical work
that compares these two discriminative sequence labeling models for
Vietnamese text processing tasks in a systematic way. We found that
the LSTMs model obtained slightly better test accuracies and had
much better memorization capacity than the CRF model in PoS tagging
and NER tasks. We also showed the effectiveness of word embedding
features in sequence labeling models. 

Because of the space limitation, we did not include maximum-entropy
Markov models (MEMM)~\cite{McCallum:2000} and hidden Markov models
(HMM)~\cite{Rabiner:1989} in our comparison. It has been shown that
the CRF model overcomes limitations of MEMM and HMM and outperforms MEMM
and HMM in many sequence labeling tasks. For the comparison between
MEMM, HMM and CRF, we can refer to the work~\cite{Lafferty:2001}. In
particular, the work~\cite{LeHong:2013a} investigated and compared
MEMM and CRF.

We also did not include bidirectional
LSTM-CNNs-CRF~\cite{ma-hovy:2016:P16-1}, the state-of-the-art
end-to-end sequence labeling model, which combine bidirectional LSTM,
CNN and CRF. That work used both word- and character-level
representations in the neural network. Actually, in our paper, we do
not aim to obtain state-of-the-art results but to compare
discriminative sequence labeling models in a basic setting.

Word representations which are learned from raw text corpora, have
been shown to be effective in sequence labeling
models. In~\cite{Turian:2010}, Turian et al. intensively evaluated
features derived from unsupervised word representations such as Brown
clusters and word vectors on NER and chunk tasks with the CRF
model. They used near state-of-the-art supervised baselines, and
showed that word representation features improved those baselines. Our
work confirmed the benefit of using word representation features for
Vietnamese language processing tasks. In this paper, although we did
not use word-cluster-based features, we obtained significant
improvements in both two tasks.

In our work, we limit the maximal length of sentences to 25
tokens. The reason is that the LSTMs model has very high computational cost,
especially for long sentences. We need about 8 hours just to train 
an experiment with the LSTMs model.\footnote{On an IBM server with 32 GB
  RAM and 8-core CPU.} Considering that the main
purpose of the paper is to compare two sequence labeling models in experiments with simple settings, we decided to limit the maximized length of
sentences. With the same reason, we decided to just use simple unigram
features in the two sequence labeling models. 

We found that the LSTMs model did not really outperform the CRF model
in our experiments. We suspect that the training data size we used in
experiments is not large, and it affected the generalization capacity
of the LSTMs model. Improving generalization capacity of deep learning
models on small data is still a challenging problem in the deep
learning research community. In contrast, the CRF model worked quite
well even with moderate-sized training data. The lesson we leaned from
the results is that in an application domain that we could not obtain
large data, we may use fast sequence labeling models such as CRF and
spend time designing good features that are specific and
beneficial for that domain. 


\section{Related Work}
\label{sec:relatedWork}

This section briefly reviews related works on Vietnamese
part-of-speech tagging and named entity recognition using
discriminative sequence models. In~\cite{Le:2010a}, the authors give
an empirical study of MEMM for Vietnamese part-of-speech tagging with
diffferent feature sets. Their best model has a tagging accuracy of
about 93.5\% when all the VLSP treebank is used. We see that despite
using a smaller data set with short sentences and a very simple
feature set with minimal hand-crafted word shapes, we are able to
achieve a tagging accuracy of more than 90\%. This is a strong
lower bound for this task when only raw text is available for tagging.

\begin{table}[h]
  \caption{Performances of NER systems at VLSP 2016}
  \label{tab:vlsp-ner}
  \begin{center}
    \begin{tabular}{|c|c|c|}
      \hline 
      \textbf{Team} & \textbf{Model} & \textbf{Performance} \\ 
      \hline
      \hline 
      Le-Hong~\cite{Phuong:2016} & ME & 88.78  \\
      \hline 
      [Anonymous] & CRF & 86.62 \\ 
      \hline 
      Nguyen et al.~\cite{Van:2016} & ME & 84.08 \\ 
      \hline
      Nguyen et al.~\cite{Son:2016} & LSTM & 83.80 \\
      \hline 
      Le et al.~\cite{Huong:2016} & CRF  & 78.40 \\ 
      \hline 
    \end{tabular}
  \end{center}
\end{table}

In VLSP 2016 workshop, several different systems have been proposed
for Vietnamese NER. The F-score of the best participating system is
88.78\%~\cite{Phuong:2016} in that shared task. That system used many
hand-crafted features to improve the performance of MEMM. Most
approaches in VLSP 2016 used the CRF and maximum entropy models, whose
performance is heavily dependent on feature
engineering. Table~\ref{tab:vlsp-ner} shows those models and their
performance. We observe that although the models studied in this work
only rely on word features, their performance is very competitive.

Most recently, a more advanced end-to-end system for
Vietnamese NER using LSTMs was proposed~\cite{Hoang:2017a}, which achieved an F1
score of 88.59\%.

\section{Conclusion}
\label{sec:conclusion}

We have presented an empirical and comparative study of two
discriminative sequence prediction models CRFs and LSTMs on two
fundamental tasks of Vietnamese text processing. We have demonstrated
the great benefit of integrating word embeddings which are trained by
an unsupervised learning method into both of the two models. These
word embeddings are able to capture semantic similarities which help
improve the prediction ability of the models, thereby increase the
part-of-speech tagging and named entity recognition accuracy by about
4.0\% and 5\%, respectively. The LSTMs model is slight better than the
CRFs model in terms of accuracy but the gap is not always significant
in moderate-sized data sets, with the cost of much longer training
time. We have also shown for the first time that a strong accuracy
lower bound for both part-of-speech tagging and named entity
recognition can be obtained by relying on only simple, word-based
features with a minimal hand-crafted features. Using a feature set of
word identities, word shapes and word embeddings, we can achieve
90.65\% of tagging performance and 86.03\% of recognition performance
on sentence not longer than 25 tokens. One practical implication of
this work is that in an application domain where large data is not
readily available, we should use fast sequence labeling models such as
CRFs and spend time designing good features that are specific and
beneficial for that domain instead of relying on complicated LSTMs
models.


\bibliographystyle{IEEEtran}
\bibliography{IEEEabrv,mybib,bibliography}

\end{document}